\documentclass[final,3p,10pt,times]{elsarticle}

\usepackage{amssymb}
\usepackage{amsmath}
\usepackage{epsfig}
\usepackage{graphicx}
\usepackage{amsmath}
\usepackage{amssymb}
\usepackage{subcaption}
\usepackage{booktabs}
\usepackage{array}
\usepackage{graphicx}
\usepackage{xcolor}
\usepackage{soul} 
\newcolumntype{L}[1]{>{\raggedright\let\newline\\\arraybackslash\hspace{0pt}}m{#1}}
\newcolumntype{C}[1]{>{\centering\let\newline\\\arraybackslash\hspace{0pt}}m{#1}}
\newcolumntype{R}[1]{>{\raggedleft\let\newline\\\arraybackslash\hspace{0pt}}m{#1}}

\usepackage{booktabs, multirow}
\usepackage{tabularx}
\def\etal{et al.}
\newcommand{\MAE}[1]{NeuroStress}
\newcommand{\Phy}[1]{PINN-Stress}

\journal{Computer Methods in Applied Mechanics and Engineering}

\begin{document}

\begin{frontmatter}

\title{Physics Informed Neural Network for Dynamic Stress Prediction}


\newcommand\authormark[1]{\textsuperscript{#1}}
\author{Hamed Bolandi\authormark{a,b*}, Gautam Sreekumar\authormark{b}, Xuyang Li\authormark{a,b}, Nizar Lajnef\authormark{a}, Vishnu Boddeti\authormark{b}\\
\vspace{3mm}
\noindent{\authormark{a} {Department of Civil and Environmental Engineering, Michigan State University, East Lansing, MI 48824}\\
\authormark{b} Department of Computer Science and Engineering, Michigan State University, East Lansing, MI 48824\\}
{* Corresponding author, Email Address: bolandih@msu.edu}}

        

\begin{abstract}
 Structural failures are often caused by catastrophic events such as earthquakes and winds. As a result, it is crucial to predict dynamic stress distributions during highly disruptive events in real time. Currently available high-fidelity methods, such as Finite Element Models (FEMs), suffer from their inherent high complexity. Therefore, to reduce computational cost while maintaining accuracy, a Physics Informed Neural Network (PINN), \Phy{} model, is proposed to predict the entire sequence of stress distribution based on Finite Element simulations using a partial differential equation (PDE) solver. Using automatic differentiation, we embed a PDE into a deep neural network's loss function to incorporate information from measurements and PDEs. The \Phy{} model can predict the sequence of stress distribution in almost real-time and can generalize better than the model without PINN.
\end{abstract}

\begin{keyword}
Physics Informed Neural Network, Stress Prediction, Finite Element Analysis, Partial Differential Equation
\end{keyword}

\end{frontmatter}

\section{Introduction}

A dynamic analysis is used to determine how a system will respond to general time-dependent loads. Events such as earthquakes and explosions are typical applications for dynamic analysis.  These applications should be able to carry out real-time analysis in the aftermath of a disaster or during extreme disruptive events that require immediate corrections to avoid catastrophic failures. Dynamic loading also can cause dramatic and damaging failures, which can be avoided by evaluating the structure during the design phase. In structural engineering, numerical analysis methods such as Finite Element Analysis (FEA)  are typically used to conduct dynamic stress analysis of various structures and systems in which it might be hard to determine an analytical solution. However, numerical methods such as FEA are computationally prohibitive while being accurate. The current workflow for FEA applications consists of: (i) modeling the geometry and its components, (ii) specifying the material properties, boundary conditions, meshing, and loading, (iii) dynamic analysis, which may be time-consuming based on the complexity of the model. The complexity of this workflow and its time requirements make it impractical for real-time applications.\\

\begin{figure*}[t]
    \centering
    \includegraphics[width=\textwidth]{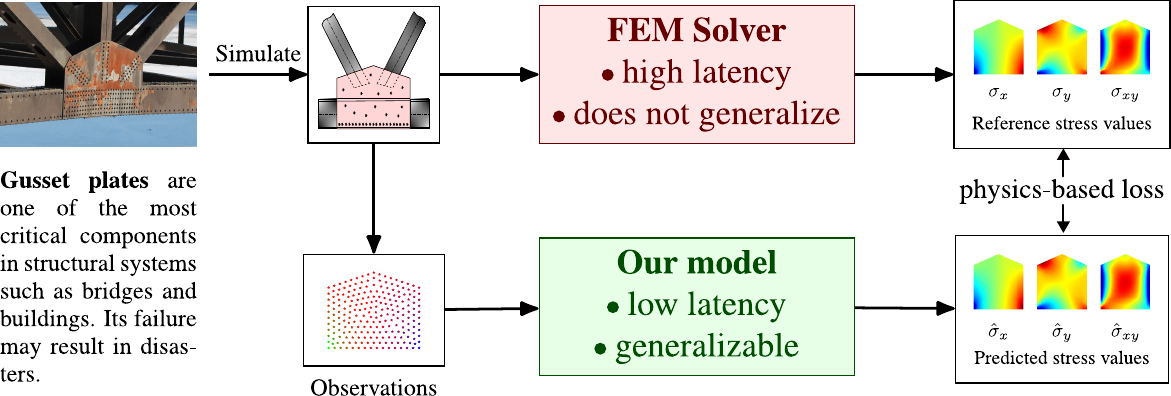}
    \setlength{\belowcaptionskip}{-10pt}
    \caption{\textbf{Overview:} Unlike FEM, \Phy{} is computationally efficient, facilitates real-time analysis and is generalizable. \Phy{} use a governing equation behind the equation of motion as a soft constraint in the loss function to enforce the loss to minimize. The points with different colors in observations correspond to the same nodes in the gusset plate. Gusset plate image is taken from ~\cite{astaneh2010gusset}}
    \label{fig:overview}
\end{figure*}

The recently introduced models~\cite{bolandi2022bridging, bolandi2022deep} were designed to predict static stress distributions using deep neural network~(DNN)-based methods in both intact and damaged structural components. The primary limitations of the above data-driven models are the incapability to produce physically consistent results and the lack of generalizability to out-of-distribution scenarios. The concept of physics-informed learning was introduced recently~\cite{raissi2017machine, raissi2018numerical, raissi2019deep} to address the computational cost of FEA and lack of generalizability to out-of-distribution scenarios.  There is special interest in Physics-informed Neural Networks (PINNs), which incorporate partial differential equations (PDEs) into the training loss function directly. However, their applications have primarily been limited to non-engineering toy simulations. Working with engineering problems such as those in structural engineering will require these models to learn several factors of variation in addition to the physical equations themselves, such as geometry. To overcome these issues, we propose a novel model for dynamic stress prediction which is real-time and generalizable and can therefore be used for stress prediction in seismic and explosions design.

We augment PINN with a novel neural architecture for predicting dynamic stress distribution to achieve fast dynamic analysis and address deficiencies of data-driven models. We model the stress distribution in gusset plates under dynamic loading to demonstrate its utility. Gusset plates are one of the most critical components in structural systems such as bridges and buildings. Since gusset plates are designed for lateral loads such as earthquakes, wind, and explosions, real-time dynamic models such as ours can help avoid catastrophic failures. In practice, the outputted stress maps from our models can be used by downstream applications for detecting anomalies such as cracks in the plates. In other words, it can act as a precursor to existing vision-based systems.

An overview of our approach is shown in Fig.~\ref{fig:overview}. To summarize our contributions, we introduce \MAE{} and \Phy{}, two novel deep learning models to learn dynamic stress distribution for complex geometries, boundary conditions, and various load sequences. Loss function in \Phy{} uses traditional MAE loss for training. \Phy{} uses the physics-informed loss function described in Section 3.1. For real-life use, our models require input from sensors placed on the plates. But since it is difficult to obtain such data for research purposes, we generate challenging synthetic data emulating dynamic stress prediction. Through extensive experiments on simulated data, we show that:

\begin{enumerate}
\item \MAE{} and \Phy{} can predict dynamic stress distribution with complex geometries, boundary conditions and various load sequences faster than traditional FEA solvers. Previous works only predict static stress distribution;

\item \MAE{} and \Phy{} can learn the temporal information in the data to make accurate predictions;

\item Introducing novel spatiotemporal multiplexing to physics-informed learning and showing its utility in dynamic stress prediction;

\item \MAE{} and \Phy{} can predict von Mises stress distribution using the von Mises equation. von Mises stress distribution is a primary diagnostic tool to predict failure of a structure;

\item To the best of our knowledge, \Phy{} is the first model that learns governing equations behind that of motion in structures. We attribute the generalization abilities of our architecture on unseen load sequences and geometries to its loss function.
\end{enumerate}

\section{Related Works}

Over the past few years, there has been a revolution in data-driven applications in various engineering fields, including fluid dynamics~\cite{farimani2017deep,kim2019deep}, molecular dynamics simulation~\cite{goh2017deep,mardt2018vampnets} and material properties prediction~\cite{mohammadi2014multigene,sarveghadi2019development,mousavi2012new,bolandi2019intelligent}. Recent studies have shown that convolutional neural networks (CNN) and Long Term Short Memories (LSTM) can be used to build metamodels for predicting time history responses.  Modares~\etal~\cite{modarres2018convolutional} studied composite materials to identify the presence and type of structural damage using CNNs. Nie~\etal~\cite{nie2020stress} developed a CNN-based method to predict the low-resolution stress field in a 2D linear cantilever beam. Jiang~\etal~\cite{jiang2021stressgan} developed a conditional generative adversarial network for predicting low-resolution static von Mises stress distribution in solid structures. Zhang~\etal~\cite{zhang2019deep} used LSTM to model nonlinear seismic responses of structures with large plastic deformations. Do~\etal~\cite{do2019fast} proposed a method for forecasting crack propagation in risk assessment of engineering structures based on LSTM and Multi-Layer Perceptron (MLP).  Yao~\etal~\cite{yao2020fea} proposed a physics-guided learning algorithm for predicting the mechanical response of materials and structures. Das~\etal~\cite{das2020data} proposed a data-driven physics-informed method for prognosis and applied it to predict cracking in a mortar cube specimen. Wang~\etal~\cite{wang2020towards} proposed a  hybrid DL model that unifies representation learning and turbulence simulation techniques using physics-informed learning. Goswami~\etal~\cite{goswami2022physics} proposed a physics-informed variational formulation of DeepONet for brittle fracture analysis. Raissi~\etal~\cite{raissi2017physics} proposed a physics-informed neural network that can  solve supervised learning tasks while respecting any given law of physics described by general nonlinear partial differential equations. Haghighat~\etal~\cite{haghighat2021physics} presented physics-informed neural networks to inversion and surrogate modeling in solid mechanics. Jin~\etal~\cite{jin2021nsfnets} investigated the ability of PINNs to directly simulate incompressible flows, ranging from laminar to turbulent flows to turbulent channel flows. Li~\etal~\cite{li2020fourier} used the Fourier transform  to develop a Fourier neural operator to model turbulent flows.

\section{Background}
\label{subsec:PhysicsLoss}

To ensure that any component of an object is in equilibrium, the balance of forces and moments acting on that component should be enforced. Stress components acting on the face of the element can be written as equations of equilibrium. The stress equilibrium equation can be written as a variation in each stress term within the body since stress changes from point to point. Considering a two-dimensional case in which stress acts in the horizontal and vertical directions gives the following set of equations of motion:

\begin{equation}\label{eq_1}
\frac{\partial \sigma_{xx}}{\partial x}+\frac{\partial \sigma_{xy}}{\partial y}+ b_x-\rho a_x=0
\end{equation}

\begin{equation}\label{eq_2}
\frac{\partial \sigma_{yy}}{\partial y}+\frac{\partial \sigma_{xy}}{\partial x}+ b_y-\rho a_y=0
\end{equation}

where $\sigma_{xx}$, $\sigma_{yy}$ and $\sigma_{xy}$ denote normal stress in horizontal and vertical directions, and shear stress respectively. $b_x$ and $b_y$ represent body force in horizontal and vertical directions. $a_x$ and $a_y$ represent an acceleration in the horizontal and vertical directions and $\rho$ denotes the density of the material.

\subsection{von Mises equation}

von Mises stress is a way of measuring whether a structure has begun to yield at any point. To compare experimentally observed yield points with calculated stresses, von Mises stress can be used mathematically as a scalar quantity. We also predict von Mises stress since the engineering community relies heavily on it. von Mises stress can be calculated from the predicted $\sigma_{xx}$, $\sigma_{yy}$, and $\sigma_{xy}$ through the von Mises stress equation.

\begin{equation}\label{eq_6}
\sigma_{vm} = \sqrt{\sigma_{xx}^2 + \sigma_{yy}^2 - \sigma_{xx} \sigma_{yy} + 3 \sigma_{yy}^2}
\end{equation}

\section{Method}

We introduce a novel architecture in this paper and augment it with a physics-based loss function for gains in generalization.

\subsection{Architecture}

Firstly, we use a 2-layered MLP to encode the input to a larger dimensional space. Then we introduce our spatiotemporal multiplexing~(STM) module to encode the spatial and the temporal information alternatively. We treat both the temporal and the spatial dimensions as sequences, which may be modeled using an appropriate deep neural architecture such as RNN, LSTM~\cite{hochreiter1997long} or self-attention~\cite{vaswani2017attention}. LSTMs have demonstrated better performance than RNNs, but have performed worse compared to self-attention. However, self-attention requires plenty of data, which cannot be satisfied in our problem statement. Hence, as a middle ground, we use LSTMs to model both temporal and spatial information.

\begin{figure*}[h]
    \centering
    \includegraphics[width=\textwidth]{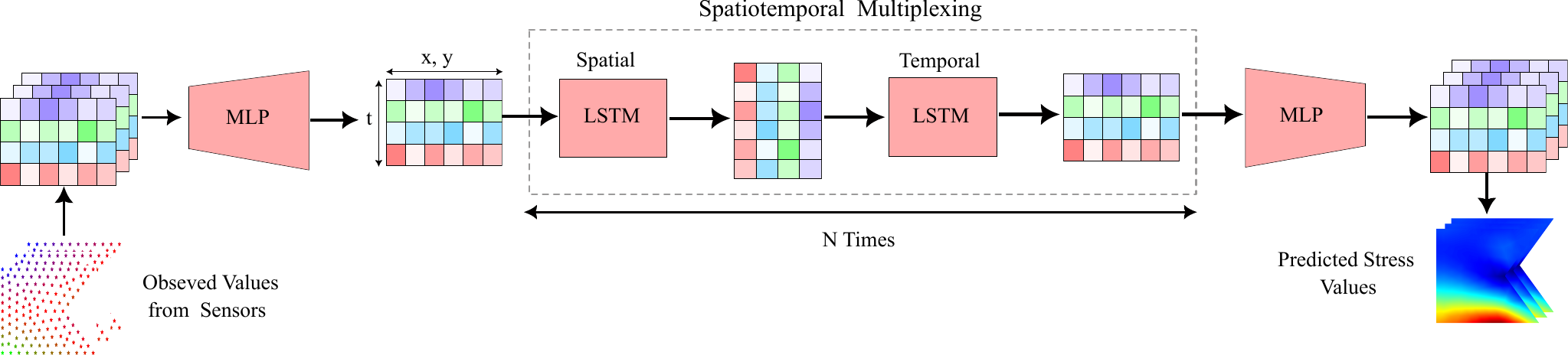}
    \caption{\textbf{Model architecture:} We introduce the novel spatiotemporal multiplexing~(STM) to physics-informed learning in order to learn both spatial and temporal information in the data. Our architecture is lightweight and hence gives real-time performance.}
    \label{fig:Network Architecture}
\end{figure*}

\textbf{Spatiotemporal multiplexing~(STM):} A single instance of our STM module consists of two LSTM layers - one for temporal sequence modeling and another for spatial sequence modeling. The input feature to an STM module is of shape $B\times N\times T\times d$ where $B, N, T, d$ are batch size, number of spatial nodes, number of time frames and feature dimension respectively. We reshape this tensor into $BN\times T\times d$ and feed it as input to the first LSTM. Here, $T$ forms the index for sequence. The output tensor from this LSTM is reshaped to $BT\times N\times d$ before feeding it into the second LSTM for spatial sequence modeling. We would like to point out that the idea of multiplexing is not novel in deep learning literature~\cite{loshchilov2017decoupled, ke2022learning}. Our contribution is that we are the first to introduce multiplexing in physics-informed learning and show its utility in dynamic prediction. Our whole architecture consists of three STM modules, totaling six LSTM layers. The architecture is schematically shown in Fig.~\ref{fig:Network Architecture}

\subsection{Physics Loss Function}

In order to force our model to learn the physical constraints, we minimize the violation of the physical equations shown in Eq.~\ref{eq_1} and~\ref{eq_2}. We also minimize the boundary condition violation to fully enforce the underlying PDE. Specifically, our loss function is a weighted sum of three loss terms:

\begin{equation}\label{eq_3}
\mathcal{L} = w_{\text{data}} \mathcal{L}_{\text{data}} + w_{\text{PDE}} \mathcal{L}_{\text{PDE}} + w_{\text{bc}} \mathcal{L}_{\text{bc}}
\end{equation}

where $\mathcal{L}_{\text{data}}$ measures the mean absolute error~(MAE) between true and predicted labels. $\mathcal{L}_{\text{PDE}}$ measures the violations of the physical equations by calculating the mean absolute error between the LHS and the RHS. $\mathcal{L}_{\text{bc}}$ corresponds to boundary condition constraints. $w_{\text{data}}$, $w_{\text{PDE}}$ and $w_{\text{bc}}$ are the weights used to balance the interplay between the three loss terms. $\mathcal{L}_{\text{bc}}$ consists of the initial and boundary conditions at each time step as below:

\begin{equation}\label{eq_4}
\sigma(x, y, t=0) = 0
\end{equation}
\begin{equation}\label{eq_5}
\sigma(x, y, (t_0...t_n)) = \sigma
\end{equation}

Equations~\ref{eq_4} and~\ref{eq_5} should be satisfied for $\sigma_{xx}$, $\sigma_{yy}$ and $\sigma_{xy}$. $x$ and $y$ are coordinates of meshes in each sample, and $t$ is the time at time steps.

\subsection{Differentiable grid from mesh}

\begin{figure}[!h]
    \centering
    \subcaptionbox{}
    {\includegraphics[scale=0.57]{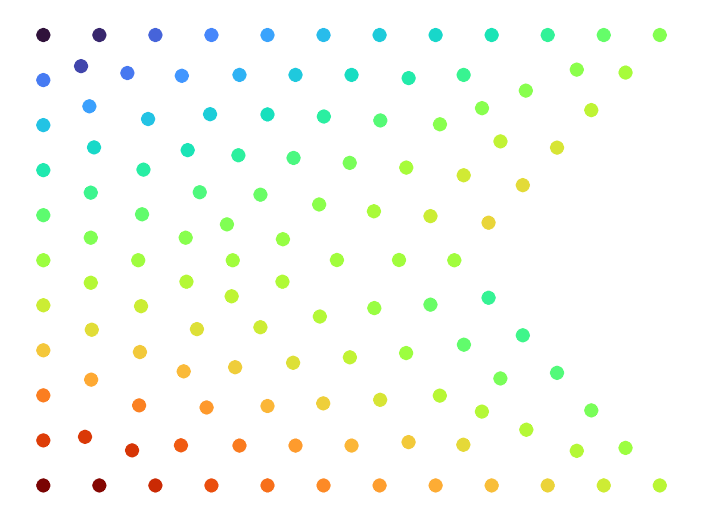}}
    \subcaptionbox{\label{fig:interpolated}}{\includegraphics[scale=0.57]{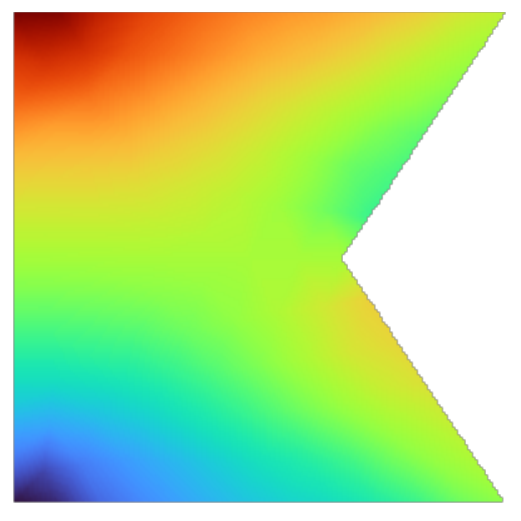}}
    \subcaptionbox{\label{fig:mesh-node-reconstructed}}{\includegraphics[scale=0.57]{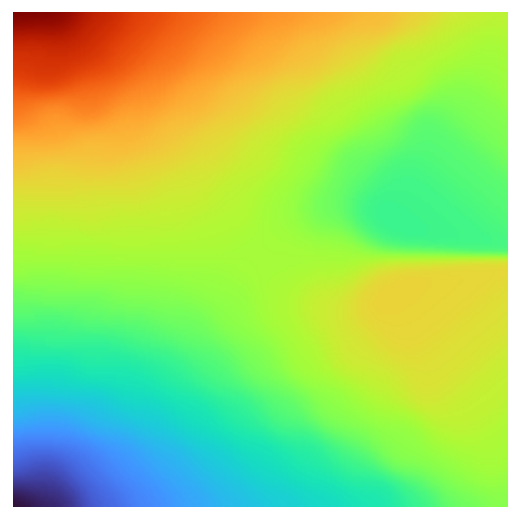}}
    \subcaptionbox{\label{fig:mask}}
    {\includegraphics[scale=0.57]{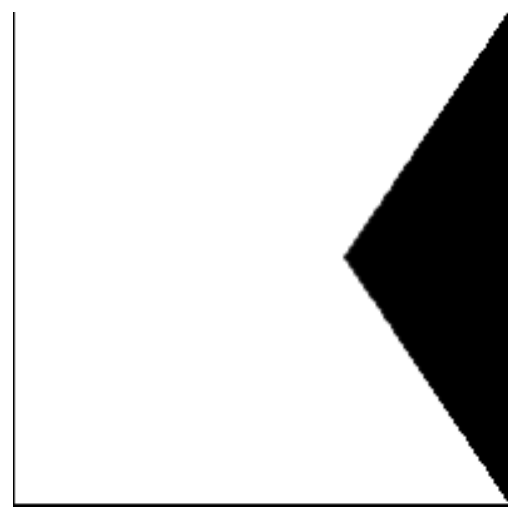}}
    \caption{\textbf{Constructing grid values from mesh values} (a) mesh nodes of a single output, the color of each node represents the stress value at the corresponding node, (b) reconstruction from Matplotlib \texttt{tricontourf} function, (c) our reconstruction on a $200\times200$ grid, (d) corresponding mask showing interpolated regions.}
    \label{fig:mesh nodes}
\end{figure}

Our physics-based loss function requires us to estimate the gradients of stress output along $x$ and $y$ directions. But since our output is in the form of a triangular mesh, gradient computation is not easy. Instead, we propose to calculate gradients on a surrogate grid created using kernel density estimation (KDE). Specifically, we calculate the stress value at a grid vertex by adding contributions from every mesh node, weighted by a Gaussian filter centered at this vertex and having a specific variance. By tuning the variance of this filter, we can achieve a robust, accurate reconstruction of the mesh along with a mask showing extrapolated regions. The original mesh, the grid reconstructed from it, and the corresponding mask are shown in Fig.~\ref{fig:mesh nodes}. To compare the accuracy of the surrogate grid, we compare it against the reconstruction obtained through \texttt{tricontourf} function in Matplotlib package in Python. As can be observed in Fig.~\ref{fig:mesh-node-reconstructed}, the grid is accurate within the mesh region. Now, we can estimate the gradients for the stress outputs from these grids.

\section{Experiments and Results}

\subsection{Data Generation}
\begin{figure}[!h]
    \centering
    \includegraphics[width=\textwidth]{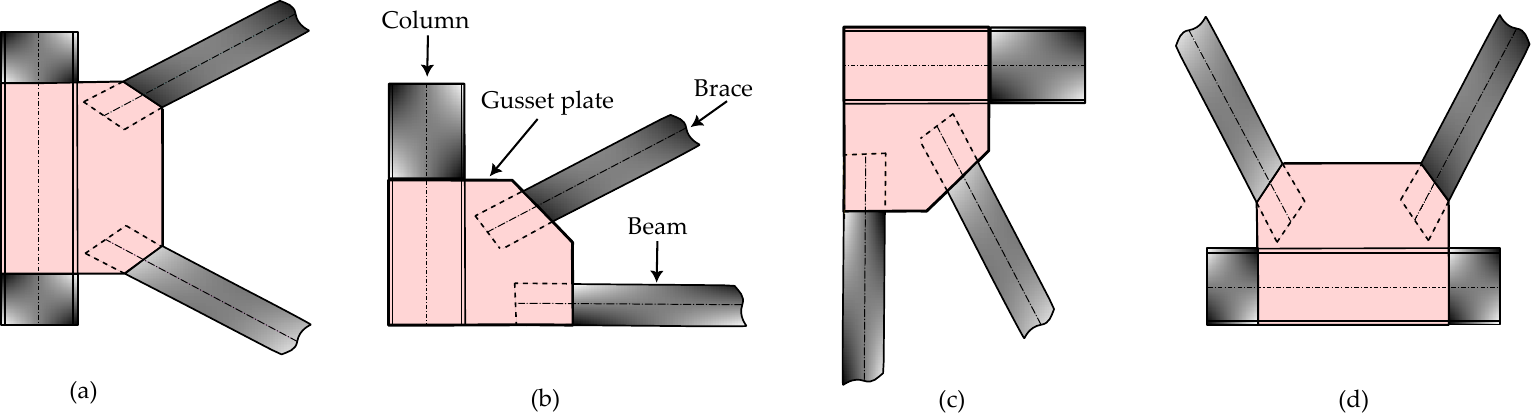}
    \caption{Some of the most common gusset plates in practice.\label{fig:common gussets}}    
\end{figure}

Gusset plates connect beams and columns to braces in steel structures. The behavior and analysis of these components are critical since various reports have observed failures of gusset plates subject to lateral loads~\cite{zahraei2007destructive,zahrai2014towards,zahrai2019numerical}. The boundary conditions and time-history load cases are considered to simulate similar conditions in common gusset plate structures under external loading. Some of the most common gusset plate configurations in practice are shown in Fig~\ref{fig:common gussets}.

\begin{figure}[!h]
    \centering
    \includegraphics[width=0.9\textwidth]{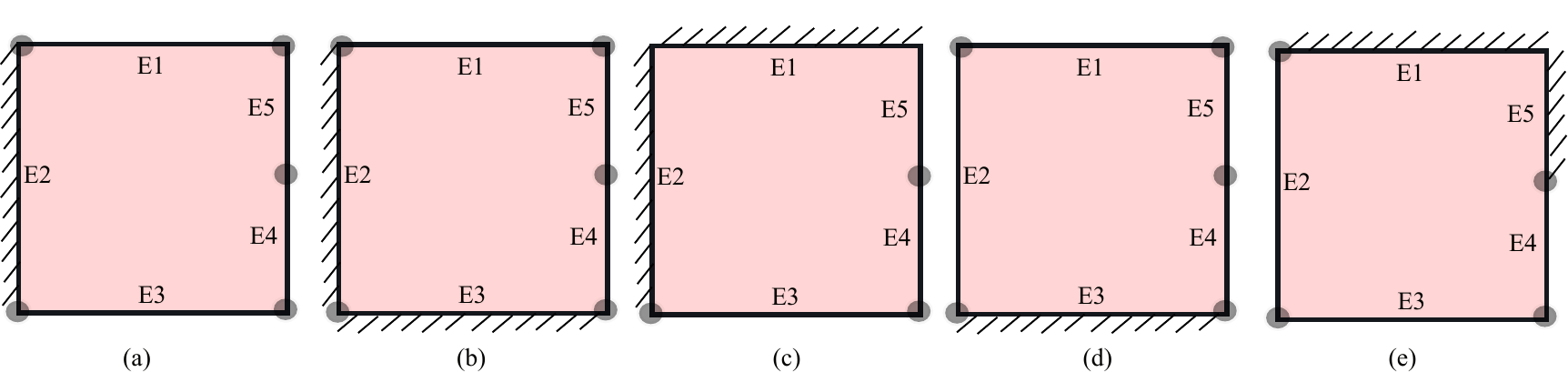}
    \caption{Different types of boundary conditions for initializing population.\label{fig:boundary conditions}}    
\end{figure}

\begin{figure}[!h]
    \centering
    \includegraphics[width=0.3\textwidth]{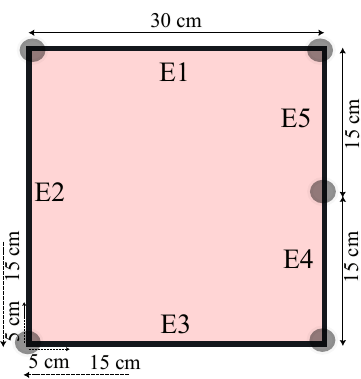}
    \caption{Basic schematic topology for initializing the steel plate geometries.}
    \label{fig: schematic topology}
\end{figure}

We create a dataset with 71,680 unique samples by combining 14 random time-history load cases, 1024 different geometries, and 5 most commonly found boundary conditions in gusset plates. Boundary conditions are shown in Fig.~\ref{fig:boundary conditions}, mimicking the real gusset plates’ boundary conditions. All the translation and rotational displacements were fixed at the boundary conditions. The range for width and height of the plates is from 30 cm to 60 cm. Two-dimensional steel plate structures with five edges, E1 to E5 denoting edges 1 to 5, as shown in Fig.~\ref{fig: schematic topology}, are considered to be made of homogeneous and isotropic linear elastic materials. Various geometries are generated by changing the position of each node in horizontal and vertical directions, as shown in Fig.~\ref{fig: schematic topology}, which leads to 1024 unique pentagons. The material properties remain unchanged and isotropic for all samples.

\begin{figure*}[t]
\centering
\subcaptionbox{\label{fig:hist1}}{\includegraphics[width=\textwidth]{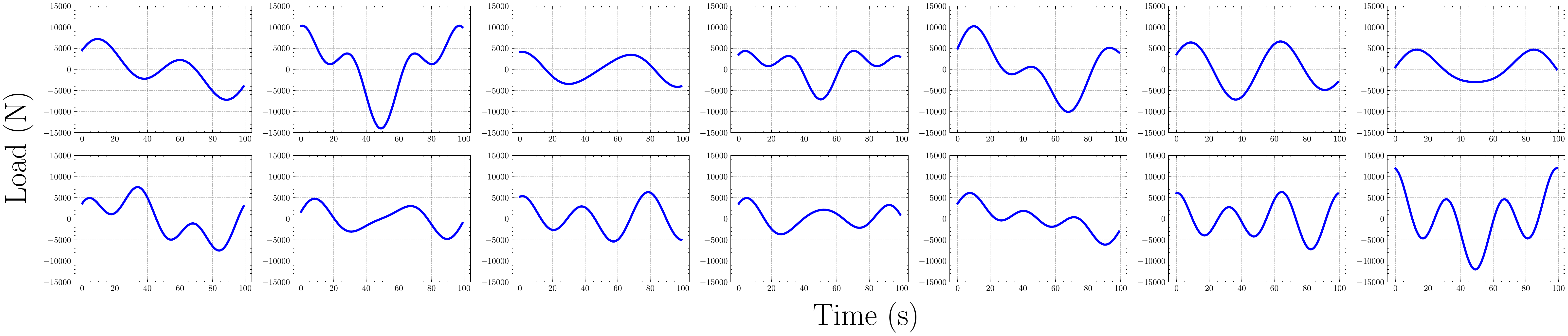}}
\subcaptionbox{\label{fig:hist2}}{\includegraphics[width=\textwidth]{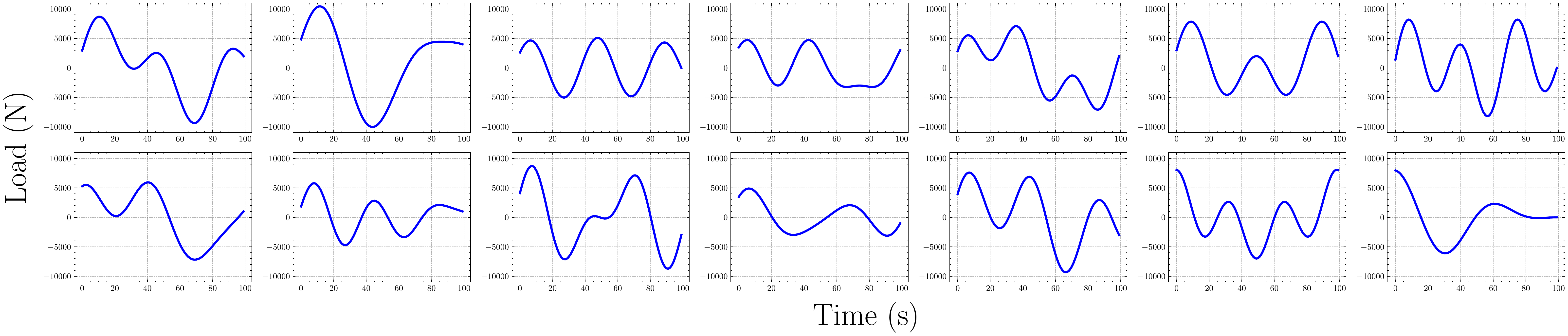}}
\caption{Various load sequences in (a) horizontal and (b) vertical directions.\label{fig:time histories}}
\end{figure*}

Time histories consist of 100 time-steps generated with random sine and cosine frequencies. The frequencies range between 1 and 3 Hz, with amplitudes ranging from 2 to 10 kN at intervals of 2 kN. All time histories in horizontal and vertical directions are shown in Fig.~\ref{fig:time histories}. Each time series last for 1 second with each time-step lasting 0.01 seconds. All the details of the input variables used to initialize train-validation-test distribution of the population are shown in Table~\ref{table:input-variables}.

\begin{table}
\begin{center}
\caption{Dataset splits}
\label{table:input-variables}
\begin{tabular}{L{0.08\textwidth} C{0.2\textwidth} C{0.18\textwidth} C{0.18\textwidth} C{0.18\textwidth}}
\toprule
 Split & Boundary condition & Load position & Load number & Geometry number \\
\midrule
 train  & E2 & E4E5 & 1-8 & 1-614  \\
train  &E2E3 & E5 & 1-8 & 1-614 \\
train  & E1E2 & E4 & 1-8 & 1-614 \\
val & E3 & E2E4 & 9-12  & 615-819\\
test  & E1E5 & E2 & 12-14 & 820-1024 \\ 
\bottomrule
\end{tabular}
\end{center}
\end{table}

\subsubsection{Input data}
\begin{figure}[!h]
    \centering
    \includegraphics[width=0.7\textwidth]{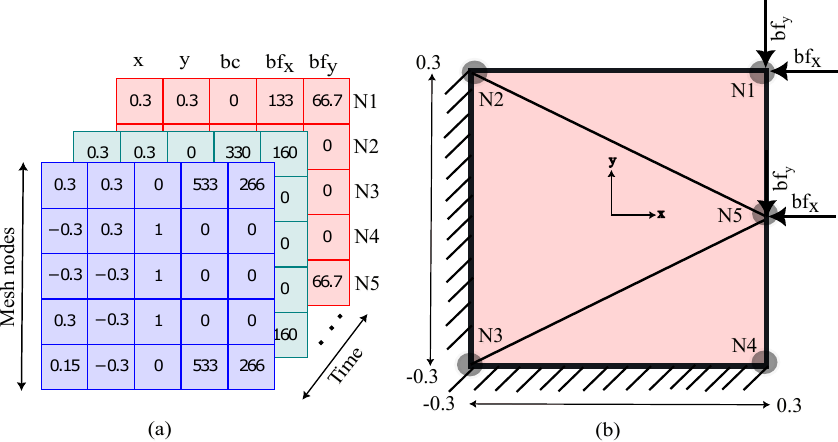}
    \caption{Construction of input matrix (Unit: m, N).\label{fig:input matrix}}    
\end{figure}

Input parameters include geometry, boundary condition, and body force in horizontal and vertical directions, each encoded as vectors in a 3-dimensional matrix. The size of the input matrix is $N\times M\times T$. where, $N$, $M$ and $T$ represent mesh nodes, input parameters and time, respectively. For example, if a sample contains 200 mesh nodes, the size of the input matrix is $200\times5\times100$. Fig.~\ref{fig:input matrix} shows how we construct the input matrix based on the geometry, boundary conditions and body forces. This figure presents a sample with five mesh nodes. However, all real samples in the trained model have more than 100 mesh nodes.  The first and the second columns of the input matrix are $x$ and $y$ coordinates of the mesh nodes respectively. The third column represents the condition of boundary constraint at each node using a Boolean value. If there is a boundary constraint at the corresponding node, then the value is one, otherwise is zero. The fourth and the fifth columns represent body force sequences at each node along $x$ and $y$ directions. Details of boundary conditions and their load positions are described in Table~\ref{table:input-variables}.

\subsubsection{Output Data}

\begin{figure}[!h]
    \centering
    \includegraphics{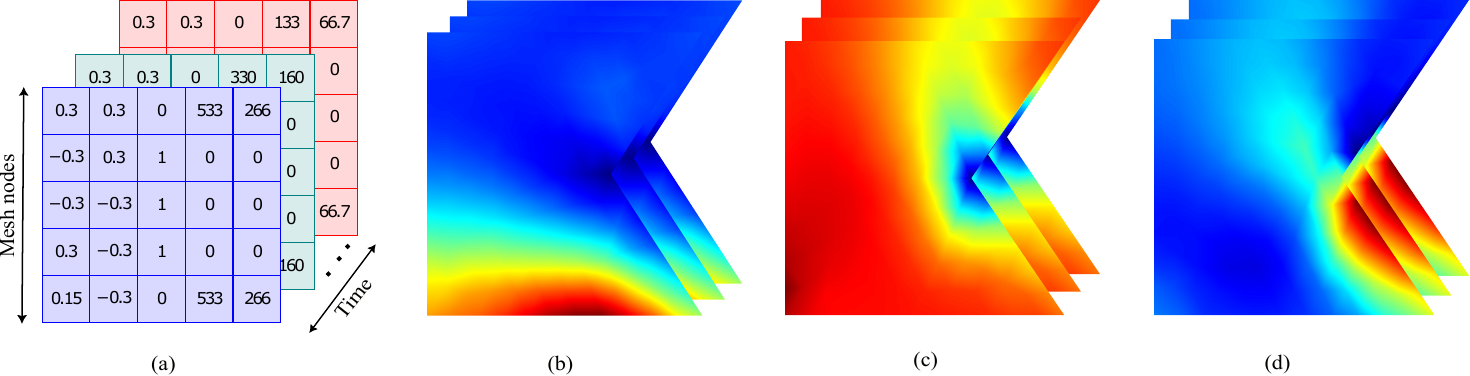}
    \setlength{\belowcaptionskip}{-10pt}
    \caption{Input and output representation for normal and shear stress distribution prediction: (a) Input matrix, (b) Output ($\sigma_{xx}$), (c) Output ($\sigma_{yy}$), (d) Output ($\sigma_{xy}$). \label{fig:Input and output}}
\end{figure}

To obtain the stress distributions for each sample, we perform FEA using the Partial Differential Equation~(PDE) solver in the MATLAB toolbox. Specifically, we use \texttt{transient-planestress} function of MATLAB PDE solver to generate dynamic stress contours which will act as the ground truth for our model. We define geometry, boundary condition, material properties, and time histories as input, and the PDE solver returns the sequence of stress distributions of $\sigma_{xx}$, $\sigma_{yy}$ and $\sigma_{xy}$ corresponding to the inputs. The size of each output is mesh nodes $\times$ load sequence. For example, if a sample contains 200 mesh nodes, the size of the output matrix is $200\times100$. Each of the three outputs are normalized separately between -1 and 1 to ensure faster convergence. The input and the output representations of the model is shown in Fig.~\ref{fig:Input and output}.

\subsection{Metrics}

We use Mean Absolute Error~(MAE), defined in Eq.~\ref{eq_6} as the primary training loss and metric. To ensure that we do not overfit to a single metric, we also use Mean Relative Percentage Error~(MRPE) to evaluate the overall quality of predicted stress distribution.

\begin{equation}\label{eq_6}
\text{MAE} = \frac{1}{NT} \sum_{N,T}^{n,t}\left|S(n,t)-\hat{S}(n,t)\right|
\end{equation}
\begin{equation}\label{eq_7}
\text{MRPE} = \frac{\text{MAE}}{ \max (|S(n,t)|, |\hat{S}(n,t)|)} \times 100
\end{equation}

\noindent where $S(n,t)$ is the true stress value at a node $n$ at time step $t$, as computed by FEA, and $\hat{S}(n,t)$ is the corresponding stress value predicted by our model, $N$ is the total number of mesh nodes in each frame of a sample, and $T$ is a total number of time steps in each sample. As mentioned earlier, we set $T=100$ in our experiments.

\subsection{Implementation}

We implemented our model using PyTorch~\cite{paszke2019pytorch} and PyTorch Lightning. AdamW optimizer~\cite{loshchilov2017decoupled} was used with an initial learning rate of $10^{-3}$. We found that a batch size of 10 gives the best results. The computational performance of the model was evaluated on an AMD EPYC 7313 16-core processor and one NVIDIA A6000 48GB GPU per experiment. The time required during the training phase for a single batch with 100 frames and a batch size of 10 for \MAE{} and \Phy{} were 7 and 20 milliseconds respectively. The inference time of \MAE{} and \Phy{} for one sample were 1 and 10 milliseconds respectively, which satisfies the real-time requirement. The most powerful FE solvers take between 10 minutes to an hour to solve the same. We use MATLAB PDE solver as a FE solver to compare the efficiency of our model. We consider the minimum time for all processes of modeling geometry, meshing, and analysis of one sample in FE solver to be about 10 minutes. MATLAB PDE solver does not use GPU acceleration. Therefore, \MAE{} and \Phy{} are about $6\times10^{5}$ and $6\times10^{4}$ times faster than MATLAB PDE solver.

\subsection{Results}

\begin{figure*}[!h]
    \centering
     \includegraphics[width=\textwidth]{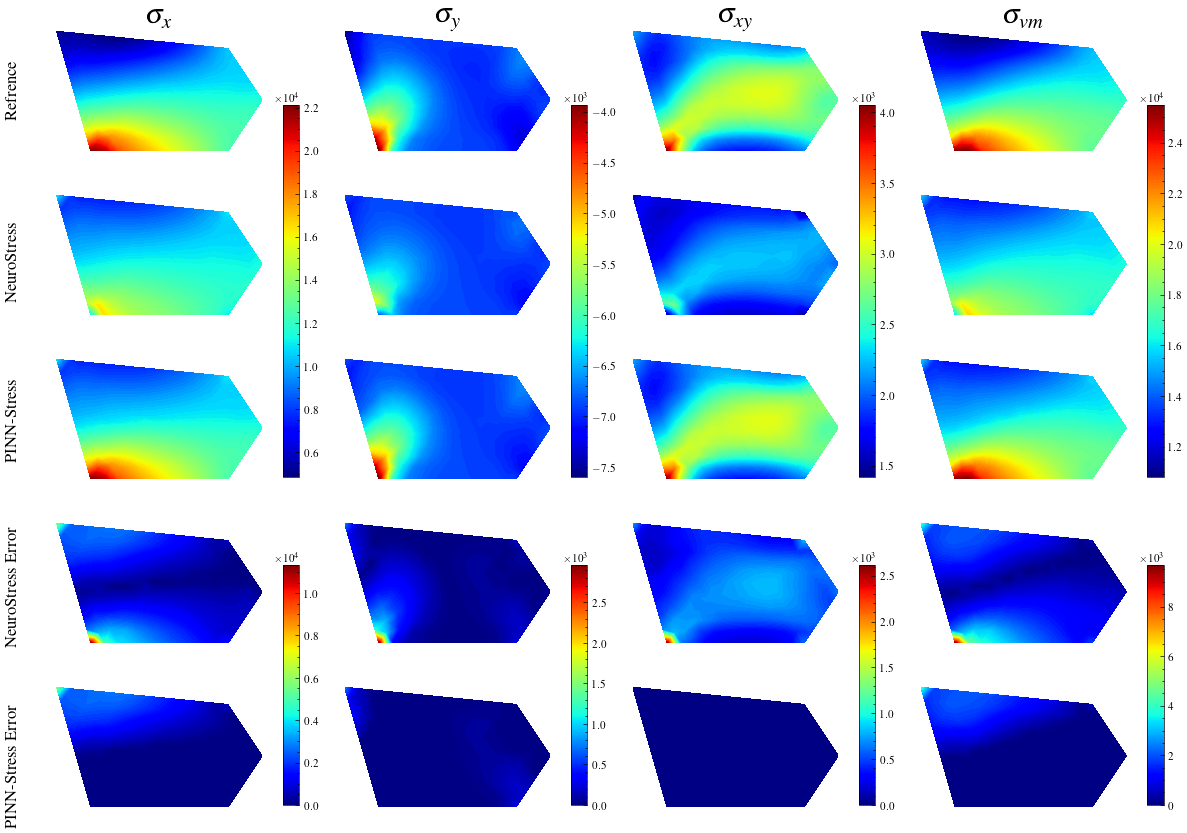}
    \caption{Comparison of \MAE{} and \Phy{} predictions for $\sigma_{xx}$, $\sigma_{yy}$, $\sigma_{xy}$ and $\sigma_{vm}$. (Unit: MPa)}
    \label{fig:MAE-Phy}
\end{figure*}

We implement two main models, \MAE{} and \Phy{}. Both models are trained on the same train dataset for 300 epochs, evaluated on the validation dataset for fine-tuning, and we report all metrics on the test dataset. The entire dataset contains 71,680 samples, while the train dataset contains 43,008 samples, validation and test datasets each contain 14336, forming the 60\%-20\%-20\% split of the whole dataset. Error metrics are calculated using the checkpoint with the least validation error. Fig.~\ref{fig:MAE-Phy} shows stress distribution prediction for $\sigma_{xx}$, $\sigma_{yy}$, $\sigma_{xy}$ and $\sigma_{vm}$ of a randomly selected frame in a sample. \Phy{} predictions are almost identical to their corresponding references, and the errors in a \Phy{} prediction are substantially lower than those in a \MAE{} prediction. Particularly, \Phy{} can capture peak stress better than \MAE{}, which is of primary importance in structural design. The importance of maximum stress matters in the design phase since maximum stress should be less than yield strength to avoid permanent deformation.

\begin{table*}[h]
\begin{center}
\caption{Data split for generalization experiments}
\label{table:generalization capability}
\begin{tabular}{L{0.13\textwidth} C{0.15\textwidth} C{0.15\textwidth} C{0.15\textwidth} C{0.15\textwidth} C{0.12\textwidth}}
\toprule
Quantity & \multicolumn{3}{c}{Data split*} & \multicolumn{2}{c}{MRPE (\%)} \\
\hline
& Train & Val & Test &\MAE{} & \Phy{} \\
\noalign{\smallskip}
\midrule
Geometry & 1-614 & 615-819 & 820-1024 & 1.7 & \textbf{1.5} \\
Load & 1-8 & 9-11 & 12-14  & 4.8 & \textbf{4.2} \\
BC & E2, E2E3, E1E2 & E3 & E1E5 & 18.3 & \textbf{16} \\
\noalign{\smallskip}
\bottomrule
\end{tabular}
\end{center}
\vspace{-1mm}
* The values in the data split column refer to indices of the corresponding generalization quantity.
\end{table*}
\vspace{-3mm}
\section{Ablation Studies}

\subsection{Generalization}

\begin{figure}[!h]
    \centering
    \includegraphics[width=0.65\textwidth]{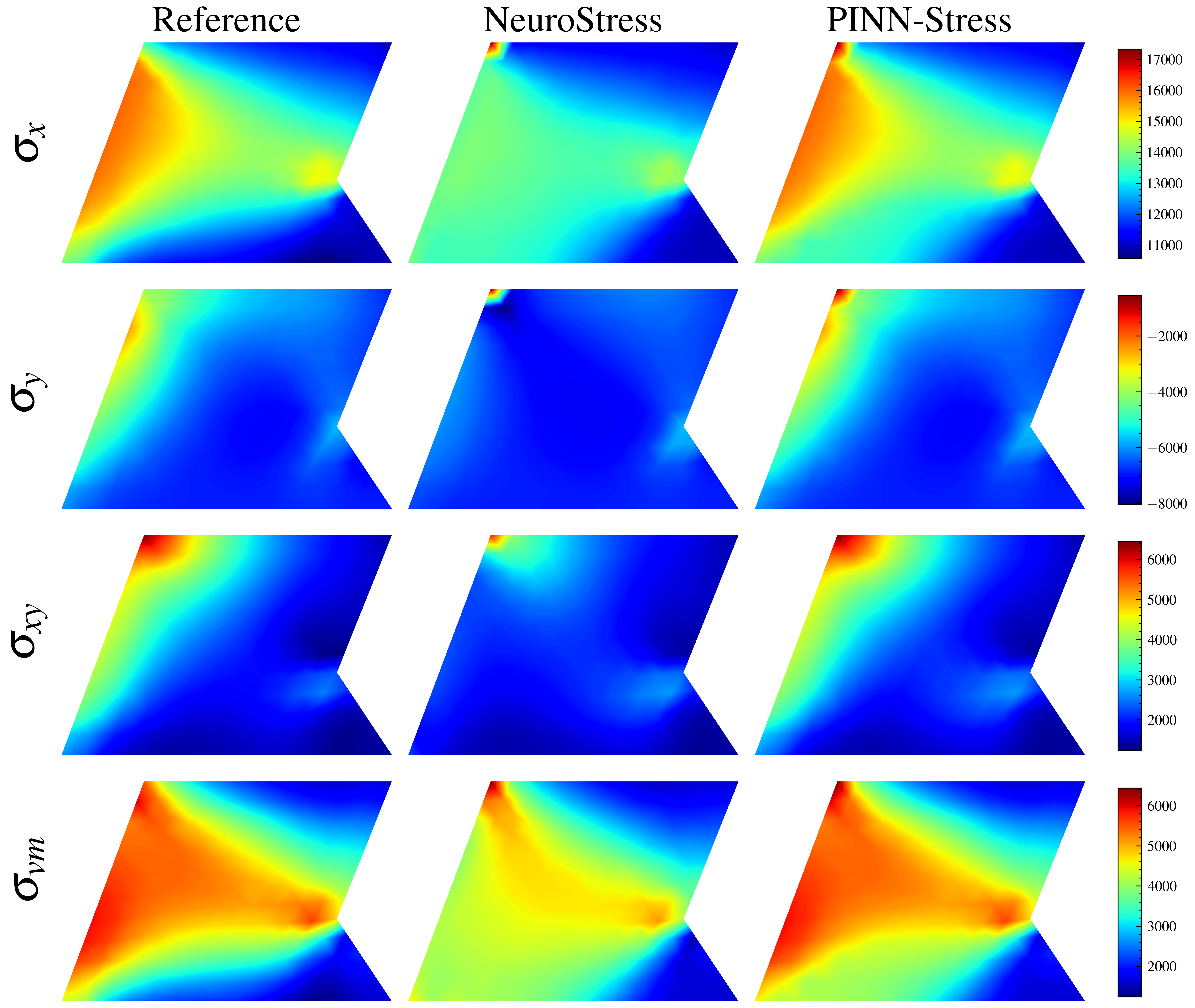}
    \setlength{\belowcaptionskip}{-10pt}
    \caption{\textbf{Predicting dynamic stress distribution for diverse load sequences:} Augmenting our novel architecture with a physics-based loss can induce generalization capabilities while still remaining real-time (Unit: MPa). The overview of our method is given in Fig.~\ref{fig:overview}.}
    \label{fig:teaser}
\end{figure}

\begin{figure}[!h]
    \centering
    \includegraphics[width=0.7\textwidth]{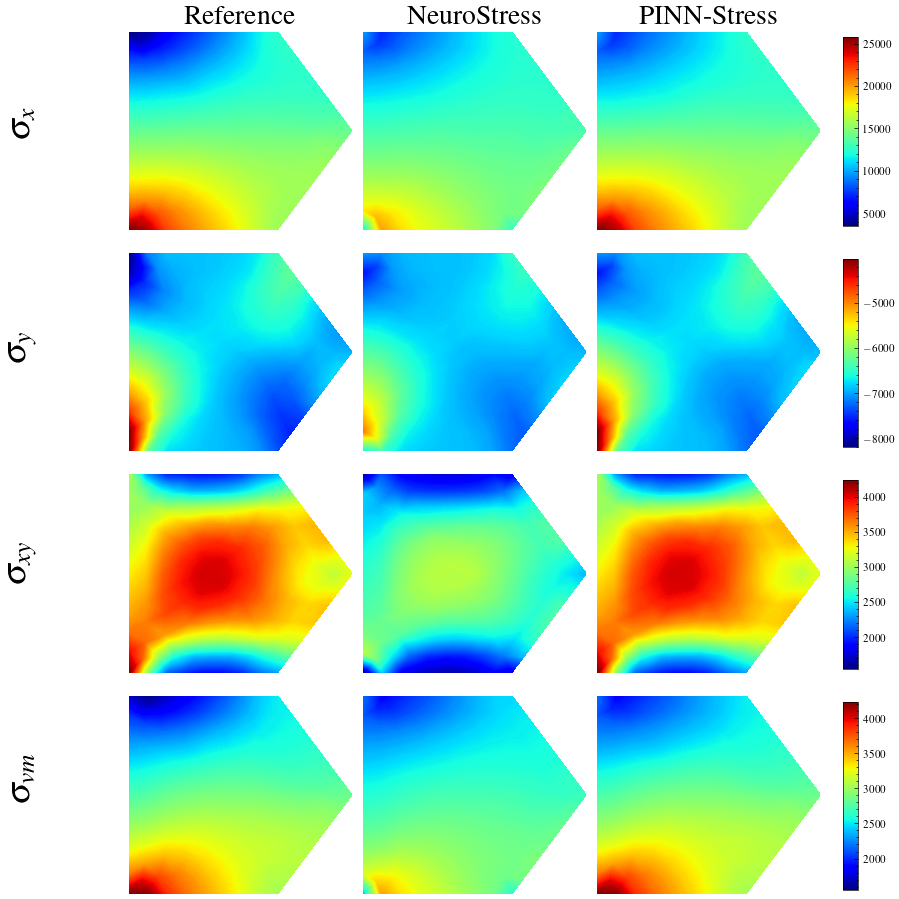}
    \setlength{\belowcaptionskip}{-10pt}
    \caption{Predicting dynamic stress distribution for diverse geometries (Unit: MPa).} 
    \label{fig:geo generalization}
    \vspace{8mm}
\end{figure}

We investigate and compare the generalization capabilities of \MAE{} and \Phy{} models for varying distributions of boundary conditions, load sequences and geometries. To that end, we collect the entire dataset and split them into train, validation and test sets such that validation and test sets contain unseen instances of the entity to check generalization on. For example, for checking generalization on geometry, train set will consist of 614 geometries out of 1024, and validation and test sets will contain the remaining (205 each). We compare the mean relative percent error~(MRPE) of each method on von Mises stress prediction. As von Mises stress identifies if a given material is likely to yield or fracture, we use its prediction error as the sole criterion. Figs~\ref{fig:teaser} and ~\ref{fig:geo generalization} demonstrates the generalization capability of \Phy{} and \MAE{} to unseen load sequences and geometries, respectively. As it can be seen, $\sigma_{xx}$, $\sigma_{yy}$, $\sigma_{xy}$ and $\sigma_{vm}$ predictions by \Phy{} are significantly better than those by \MAE{}.

\begin{figure*}[!h]
\centering
\subcaptionbox{\label{fig:per-pixel-load}}{\includegraphics[width=0.48\textwidth]{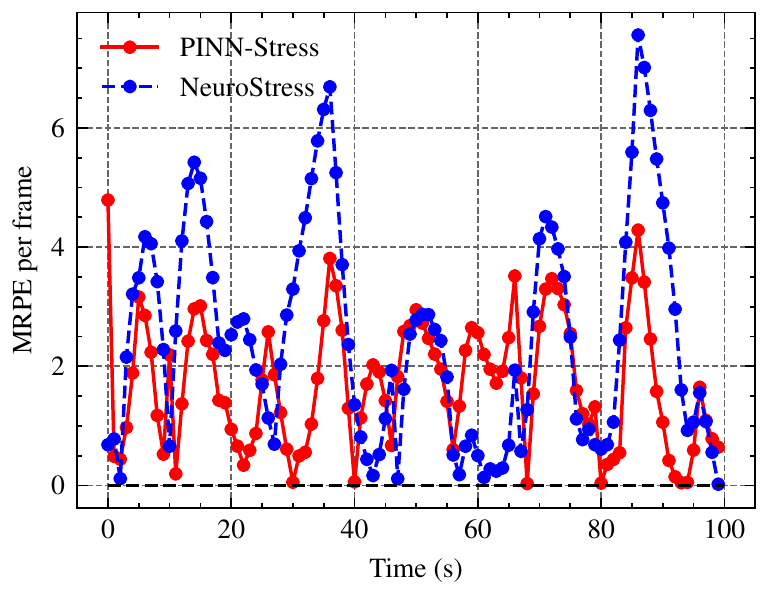}}
\subcaptionbox{\label{fig:per-pixel-geo}}{\includegraphics[width=0.48\textwidth]{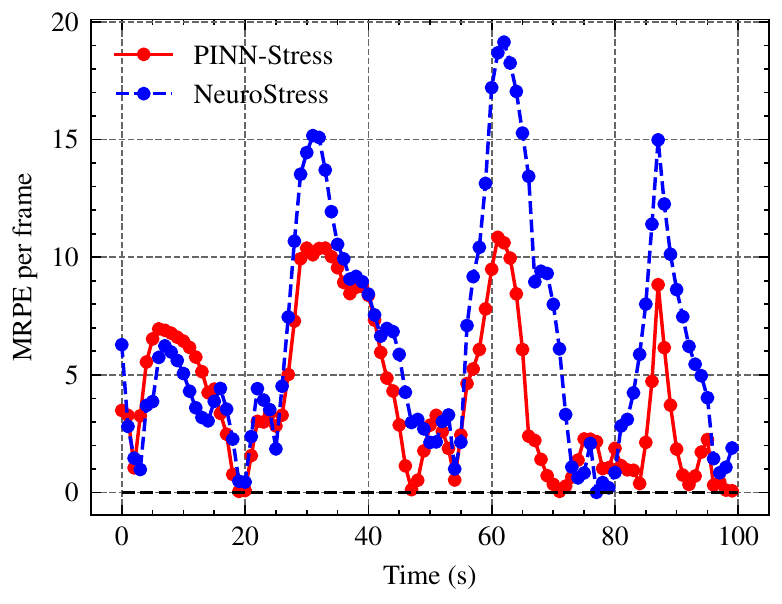}}
\caption{Comparison of \MAE{} and \Phy{} errors for $\sigma_{vm}$ across 100 frames for a random spatial node in a sample. (a) unseen load sequences and (b) unseen geometries.\label{fig:error comparision}}
\end{figure*}

\begin{figure}[!h]
    \centering
    \includegraphics[width=0.99\textwidth]{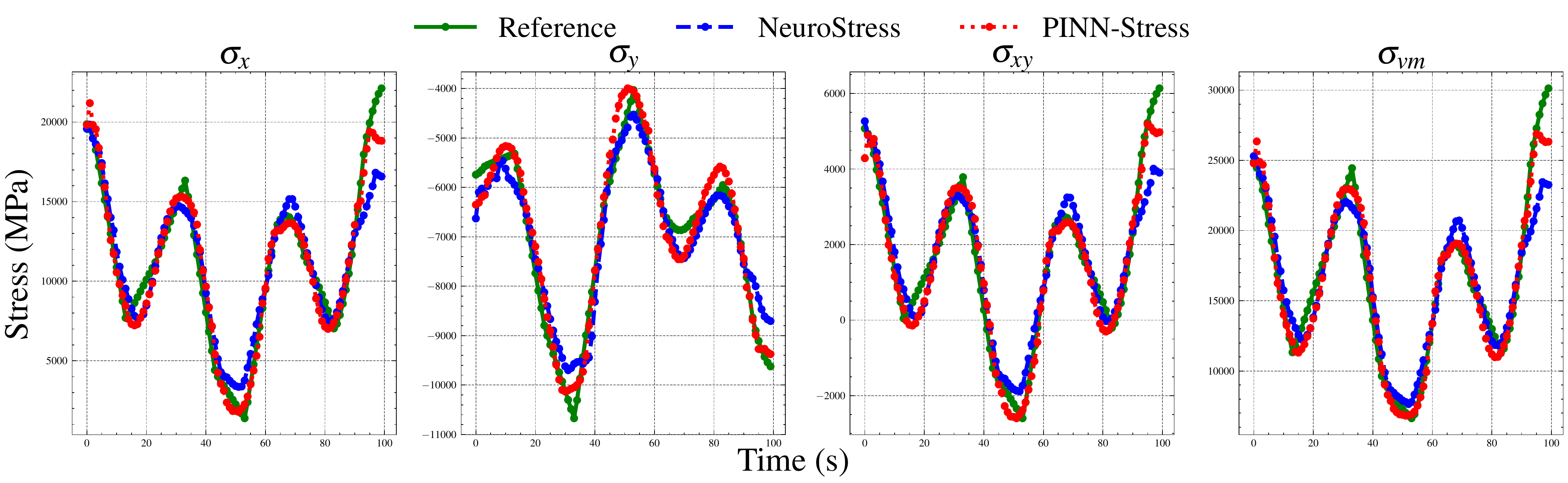}
    \caption{Comparison of \MAE{} and \Phy{} predictions for $\sigma_{xx}$, $\sigma_{yy}$, $\sigma_{xy}$ and $\sigma_{vm}$ across 100 frames for a sample with unseen load sequences.}
    \label{fig:one-element}
\end{figure}

\begin{figure}[!h]
    \centering
    \includegraphics[width=.99\textwidth]{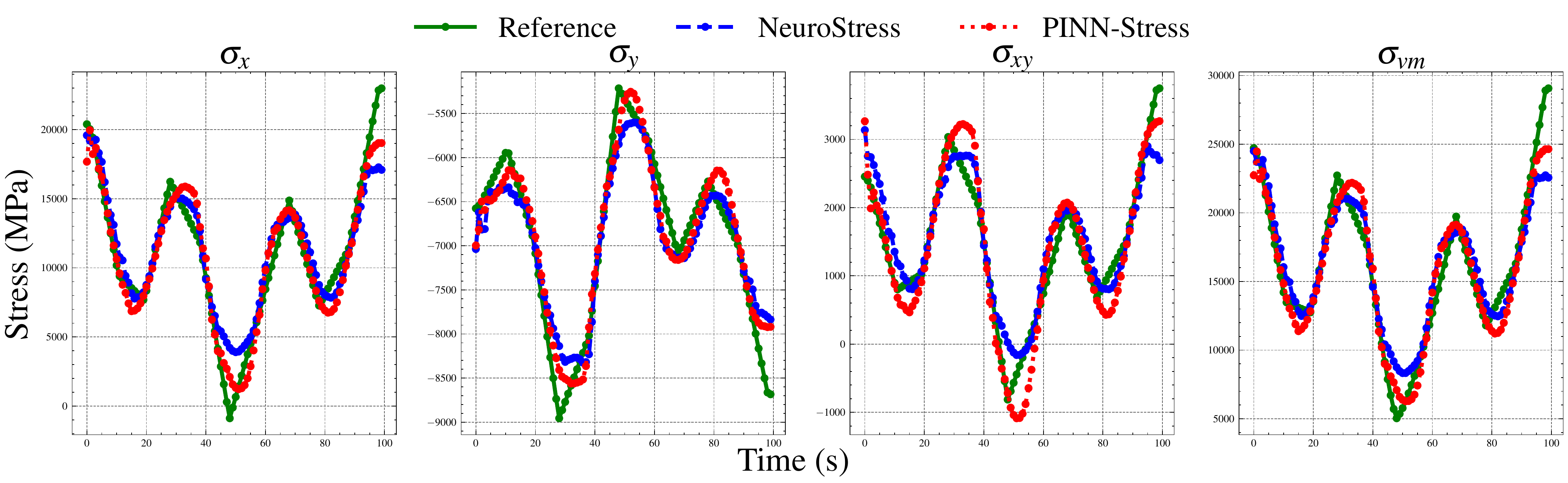}
    \caption{Comparison of \MAE{} and \Phy{} predictions for $\sigma_{xx}$, $\sigma_{yy}$, $\sigma_{xy}$ and $\sigma_{vm}$ across 100 frames for a sample with unseen geometries.}
    \label{fig:one-element-geo}
\end{figure}

Fig~\ref{fig:error comparision} shows the error of each frame for a random spatial node across all time frames for unseen load sequences and structural geometries. As it can be seen, in both figures, the errors in \Phy{} are less than \MAE{}, especially in extreme peaks, which demonstrates the ability of \Phy{} to predict the maximum stress values. We have also compared the generalization capability of \Phy{} and \MAE{} over unseen load sequences and geometries in a single spatial node across all time frames in Figs~\ref{fig:one-element} and \ref{fig:one-element-geo}.  Figs~\ref{fig:one-element} and \ref{fig:one-element-geo} demonstrate the ability of our models to capture the temporal dependencies over time frames. It can be seen that both models' predictions are almost identical to references in all the time frames. However, in extreme peaks \Phy{} outperforms \MAE{}. Table~\ref{table:generalization capability} shows the data split for each experiment and the corresponding results. The lowest error in each experiment is highlighted in bold.

In every experiment, we can observe that \Phy{} generalizes better than \MAE{}. However, neither method generalizes satisfyingly for various boundary conditions. Since we only considered five different boundary conditions in total, we ran the same experiment for different combinations of boundary conditions. The results were similar.

\subsection{Choice of architecture}

The efficiency of architecture can be attributed to several design choices we have made. Our architecture models the temporal dependency between time frames and the relationship between different nodes in an input via our spatiotemporal multiplexing mechanism. As mentioned earlier, we are the first to introduce such a design into PINNs to the best of our knowledge. Even though self-attention has shown state-of-the-art performance in sequence modeling, they are not suitable for tasks without large amounts of data. Hence, we use LSTMs for sequence modeling. To demonstrate our claim, we compare our architecture against other baseline architectures.\\

We compare against three architectures: \textbf{Spatiotempo-Att}, \textbf{Tempo-LSTM}, \textbf{Spatio-MLP}. Spatiotempo-Att is very similar to our architecture, except the LSTM modules in our model are replaced with self-attention modules. Tempo-LSTM is also similar to our architecture except the LSTMs act only along the temporal dimension. Spatio-MLP is a normal feedforward network with six layers with LeakyReLU activation in between. It treats each time frame separately but considers all the nodes simultaneously. We will refer to our architecture as \textbf{Spatiotempo-LSTM}. To save time and resources, we train all the architectures on 10\% of training data with MAE loss. Similar to our experiments on generalization, we report the error on von Mises stress prediction. The results are shown in Table~\ref{table:arch-comparison}, and the best results are highlighted in bold.


\begin{table}[!h]
\begin{center}
\caption{Architecture comparison}
\label{table:arch-comparison}
\begin{tabular}{l c c c c c}
\toprule
\multicolumn{5}{c}{Architecture} \\
\hline
 & Spatiotempo-Att & Tempo-LSTM & Spatio-MLP  & Spatiotempo-LSTM \\
\noalign{\smallskip}
\midrule
 \#Params ($K$) & 309 & 208 & 828 & 208 \\
 MRPE(\%) & 19.5 & 17.5 & 25.4 & \textbf{16.6} \\
\noalign{\smallskip}
\bottomrule
\end{tabular}
\end{center}
\end{table}

\section{Conclusion}

We propose \MAE{} and \Phy{}, two models for dynamic stress prediction based on a novel architecture, with the latter augmented with physics-informed loss function. Our models explicitly learn both spatial and temporal information through our spatiotemporal multiplexing~(STM) module. Experiments on simulated gusset plates show that not only are our models accurate, but adding physics-informed loss function facilitates generalization with respect to varying load sequences and structural geometries. \Phy{} is also better at estimating high stress values which is of more importance to the structural engineering community. However, collecting sufficient data points from real gusset plates using sensors can be expensive and noisy. Therefore, our future efforts will be directed towards achieving lower sample complexity under noisy conditions.

\section{Acknowledgment}
This research was funded in part by National Science Foundation, United States of America, grant number CNS 1645783. The corresponding author states that there is no conflict of interest on behalf of all authors.

{\small
\bibliographystyle{elsarticle-num-names}
\bibliography{main}
}
\end{document}